\documentclass[letterpaper, 10pt, conference]{ieeeconf}
\IEEEoverridecommandlockouts  
\usepackage{amsmath,amssymb,amsfonts}
\usepackage{graphicx}
\usepackage{textcomp}
\usepackage{xcolor}
\usepackage{caption}
\usepackage{subcaption}
\usepackage{multirow}
\usepackage[inkscapelatex=false]{svg}
\usepackage{algorithm}
\usepackage{algpseudocode}
\usepackage{tikz}
\usepackage{pgfplots}
\usepackage{pdfpages}
\usepackage{makecell}

\newcommand{\msefull}{$MSE_{acc}$}
\newcommand{\mse}{$MSE_{acc}95$}

\def\BibTeX{{\rm B\kern-.05em{\sc i\kern-.025em b}\kern-.08em
    T\kern-.1667em\lower.7ex\hbox{E}\kern-.125emX}}
\begin{document}

\title{\textbf{Diffusion-Based Generation and Imputation of Driving Scenarios from Limited Vehicle CAN Data}\\
\thanks{\textsuperscript{1} J. Ripper, M. Mühlhäuser and T. Kreutz are with Telecooperation Lab at Technical University of Darmstadt, Darmstadt, Germany, \{julian.ripper@stud, kreutz@tk, max@tk\}.tu-darmstadt.de\\
\textsuperscript{2} O. Esbel and R. Fietzek are with Compredict GmbH, Darmstadt, Germany, \{esbel, fietzek\}@compredict.de}
}
\author{Julian Ripper\textsuperscript{1}
\and
Ousama Esbel\textsuperscript{2}
\and
Rafael Fietzek\textsuperscript{2}
\and
Max Mühlhäuser\textsuperscript{1}
\and
Thomas Kreutz\textsuperscript{1}
}

\maketitle

\begin{abstract}
Training deep learning methods on small time series datasets that also include corrupted samples is challenging. Diffusion models have shown to be effective to generate realistic and synthetic data, and correct corrupted samples through imputation. In this context, this paper focuses on generating synthetic yet realistic samples of automotive time series data. We show that denoising diffusion probabilistic models (DDPMs) can effectively solve this task by applying them to a challenging vehicle CAN-dataset with long-term data and a limited number of samples. Therefore, we propose a hybrid generative approach that combines autoregressive and non-autoregressive techniques. We evaluate our approach with two recently proposed DDPM architectures for time series generation, for which we propose several improvements. To evaluate the generated samples, we propose three metrics that quantify physical correctness and test track adherence. Our best model is able to outperform even the training data in terms of physical correctness, while showing plausible driving behavior. Finally, we use our best model to successfully impute physically implausible regions in the training data, thereby improving the data quality.
\end{abstract}

\newcommand{\tabCar}{
\begin{table}
\vspace{0.5em}
\caption{Results for different training configurations. $MSE$ speed and $MSE$ swa denotes the $MSE$ between the true future window and the generated one for the respective channel. All evaluations are performed on 4096 samples. Ablation exp. denotes the SSSD\textsuperscript{S4} ablation experiment.}
\centering
\resizebox{\columnwidth}{!}{%
\begin{tabular}{| l |c c c c|}
\hline
\centering \thead{Training \\ configs} & \thead{signs \\ scores} $\uparrow$ & \textbf{MSE\textsubscript{acc}95} $\downarrow$ & \thead{MSE \\ speed/swa} $\downarrow$ & \thead{Difference \\ between \\ windows} $\downarrow$ \\ \hline
LongConv & 0.888 & 0.055 & 6.320 / 4.964 & 0.069 \\
\hline
SSSD\textsuperscript{S4} & 0.890 & 0.036 & 3.434 / 2.531 & 0.046 \\
\ Ablation exp. & 0.910 & 0.040 & 5.398 / 3.240 & 0.058 \\
\hline
SSSD\textsuperscript{Mamba} & 0.878 & 0.036 & 2.652 / 2.478 & 0.040 \\
\hline
\hline
Training data & 0.910 & 0.084 & & \\
\hline
\end{tabular}
}
\label{tab:Car}
\end{table}
}

\newcommand{\tabCarLap}{
\begin{table}
\caption{Results for different training configurations when generating a lap. TAM speed/swa denotes the mean TAM for the respective channel for the 16 full laps.}
\centering
\begin{tabular}{| l |c c c|}
\hline
\thead{Training \\ configs} & \thead{signs \\ scores} $\uparrow$ & \textbf{MSE\textsubscript{acc}95} $\downarrow$ & \thead{TAM \\ speed/swa} $\downarrow$ \\ \hline
SSSD\textsuperscript{S4} & 0.934 & 0.014 & 12.878 / 4.593 \\
SSSD\textsuperscript{Mamba} & 0.937 & 0.021 & \ 3.589 / 1.631 \\
\hline
\end{tabular}
\label{tab:CarLap}
\end{table}
}

\newcommand{\imputation}{
\begin{table*}
\vspace{0.5em}
\caption{Evaluation of the imputation experiments. For all settings, we show the \msefull{} and the quantified continuity for each channel and the start (left) and end (right) of the imputed regions separately.}
\centering
\begin{tabular}{| l |c |c ||c |c c |c c |c c|}
\hline
\textbf{Schedule} & \textbf{Imputed channels} & \textbf{Steps} & \textbf{MSE\textsubscript{acc}} $\downarrow$ & \multicolumn{2}{|c}{\textbf{Speed} $\downarrow$} & \multicolumn{2}{|c}{\textbf{Torques} $\downarrow$} & \multicolumn{2}{|c|}{\textbf{SWA} $\downarrow$} \\ \hline
\multirow{4}{*}{Naive} & \multirow{2}{*}{All channels}  & 16 & 0.043 & 0.343 & 2.000 & 9.573 & 20.879 & 0.320 & 0.641 \\
 &  & 500 & 0.042 & 0.295 & 2.037 & 10.571 & 14.617 & 0.224 & 0.415 \\
 & \multirow{2}{*}{Torques} & 16 & 0.041 & 0 & 0 & 10.509 & 18.845 & 0 & 0 \\
 &  & 500 & 0.042 & 0 & 0 & 9.868 & 17.572 & 0 & 0 \\
\hline
\multirow{4}{*}{RePaint} & \multirow{2}{*}{All channels} & 64 & 0.054 & 0.393 & 2.445 & 7.820 & 11.229 & 0.178 &  0.406 \\
 &  & 96 & 0.053 & 0.377 & 2.396 & 8.079 & 10.029 & 0.157 & 0.333 \\
 & \multirow{2}{*}{Torques} & 64 & 0.048 & 0 & 0 & 7.637 & 9.282 & 0 & 0 \\
 &  & 96 & 0.044 & 0 & 0 & 7.341 & 9.252 & 0 & 0 \\
 \hline
\end{tabular}
\label{tab:Imputation}
\end{table*}
}

\newcommand{\figMSEgood}{
\begin{figure}
     \centering
     \includegraphics[width=\columnwidth]{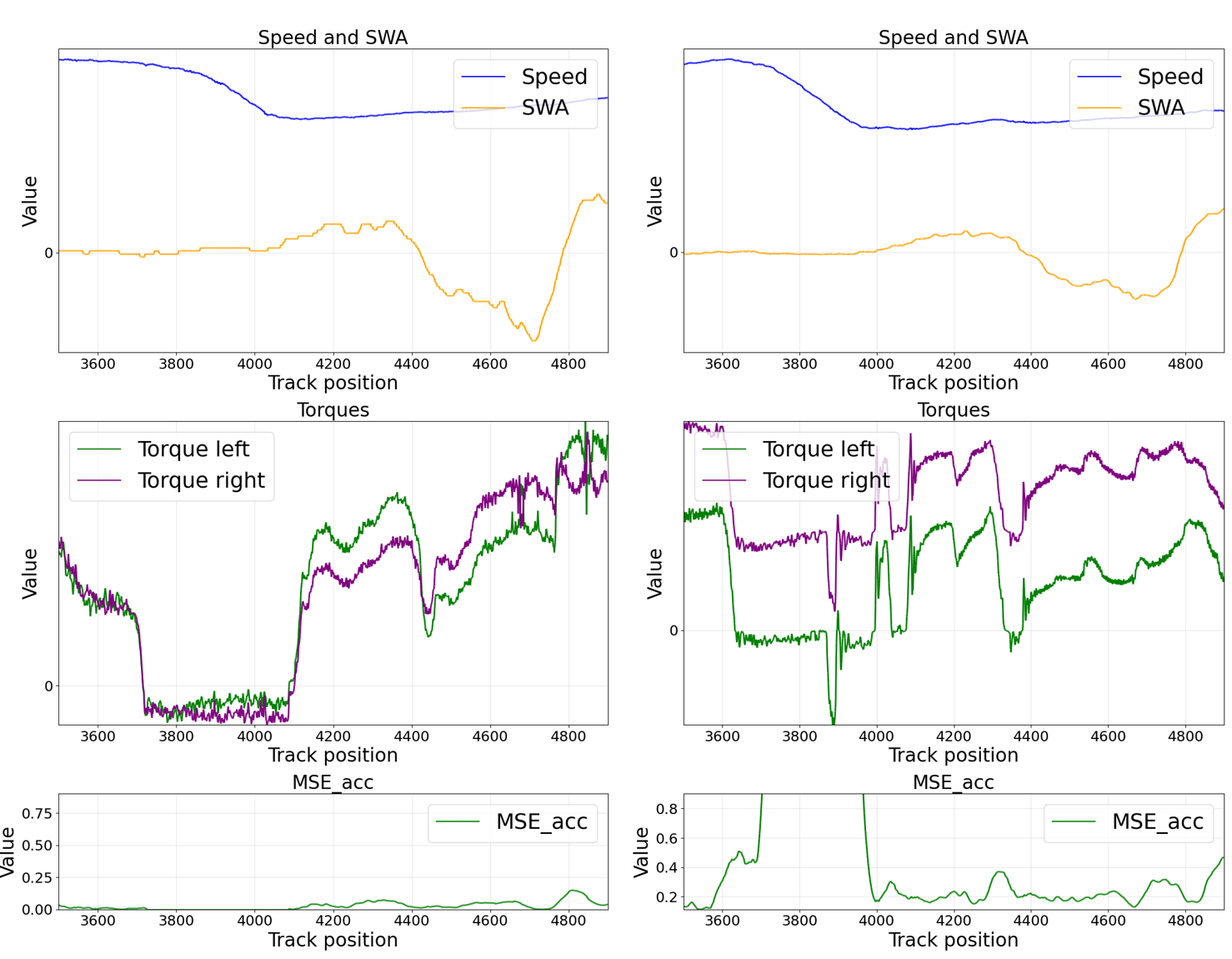}
        \caption{Qualitative evaluation of two crops of the training data, to validate the \msefull{} measure. On the left is a crop, where the sensors correctly calibrated and the \msefull{} is low. On the right the sensors are wrongly calibrated. The \msefull{} is high, especially in regions where the car is breaking, but the average of the torques is still positive.}
        \label{fig:MSEgood}
\end{figure}
}

\newcommand{\figGenTrack}{
\begin{figure}
     \centering
         \includegraphics[width=\columnwidth]{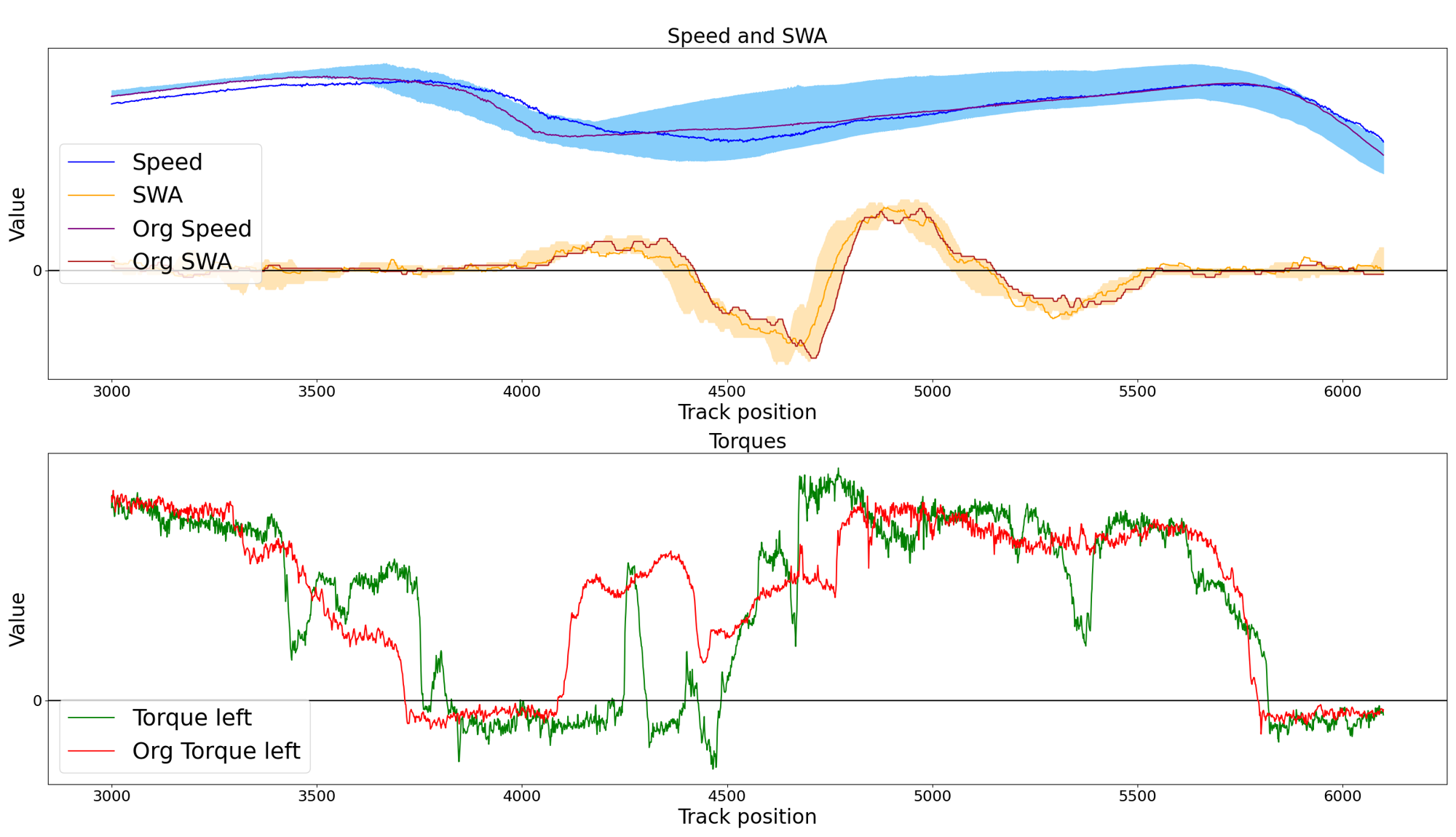}
        \caption{A crop of a lap generated by our SSSD\textsuperscript{Mamba} model. The light blue and orange regions represent the realistic range of the TAM metric for speed and swa, respectively.}
        \label{fig:LabGen}
\end{figure}
}

\newcommand{\figApproach}{
\begin{figure}
\centering
\includegraphics[width=\columnwidth]{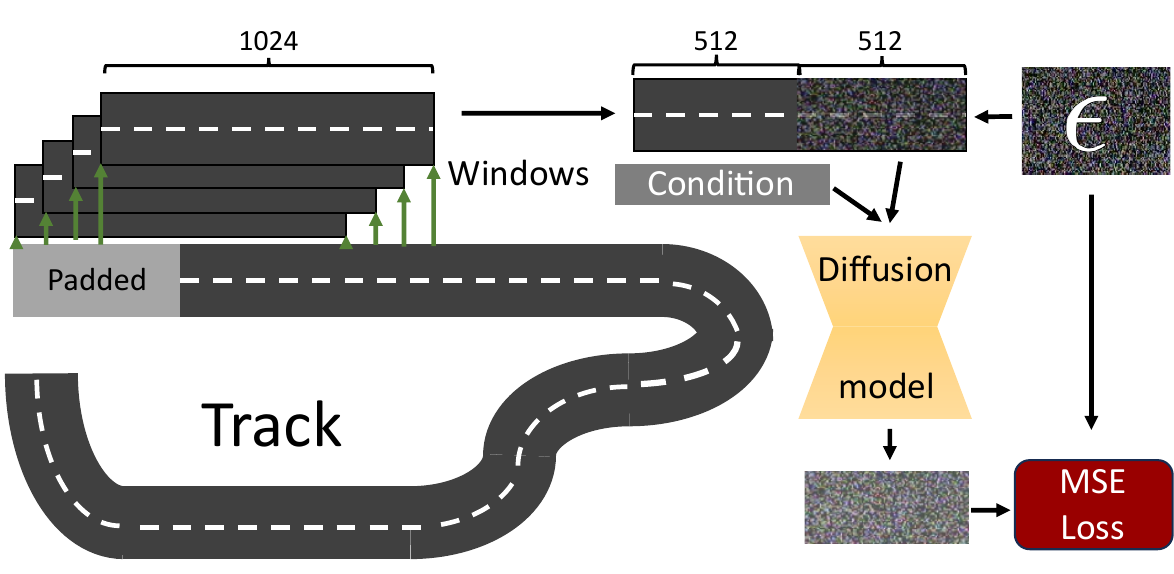}
\caption{Overview of our approach. We segment the padded track into windows, add noise to the later half, and train our diffusion model to predict the noise.}
\label{fig:approach}
\end{figure}
}

\newcommand{\MyAlgorithm}{
\begin{algorithm}[h]
\caption{Autoregressive Lap Generation Process}
\label{alg:gen}
\begin{algorithmic}[1]
\Require Generator $\mathcal{G}_\theta$, quality metric $\mathcal{M}$, initial window $W_{0,\text{past}}$, contextual features $C$
\State $X_{\text{syn}} \gets [W_{0,\text{past}}]$
\State $i \gets 0$
\While{lap not complete}
    \For{$j = 1$ to $b$}
        \State $\hat{W}_{i,\text{future}}^{(j)} \gets \mathcal{G}_\theta(W_{i,\text{past}}, C_i, j)$
    \EndFor
    \State $j^* \gets \arg\min_{j \in \{1,2,...,b\}} \mathcal{M}(\hat{W}_{i,\text{future}}^{(j)})$
    \State Append $\hat{W}_{i,\text{future}}^{(j^*)}$ to $X_{\text{syn}}$
    \State $W_{i+1,\text{past}} \gets \hat{W}_{i,\text{future}}^{(j^*)}$
    \State $i \gets i + 512$
\EndWhile
\State \Return $X_{\text{syn}}$
\end{algorithmic}
\end{algorithm}
}

\newcommand{\TimeStepsEvalSFourMSE}{
(2, 216.79617) (4, 0.35908) (6, 0.04583) (8, 0.03897) (12, 0.03439) (16, 0.03408) (24, 0.03593) (32, 0.03419) (48, 0.03430) (64, 0.03292) (96, 0.03466) (128, 0.03439) (192, 0.03420) (256, 0.03498) (384, 0.03476) (500, 0.03513)
}
\newcommand{\TimeStepsEvalSFourVel}{
(2, 100000) (4, 11.35156) (6, 4.74219) (8, 3.87891) (12, 3.62500) (16, 3.63672) (24, 3.72070) (32, 3.65234) (48, 3.70117) (64, 3.67383) (96, 3.79102) (128, 3.68945) (192, 3.72656) (256, 3.73438) (384, 3.71484) (500, 3.75391)
}
\newcommand{\TimeStepsEvalSFourSwa}{
(2, 100000) (4, 27.67188) (6, 5.32812) (8, 3.19531) (12, 2.70312) (16, 2.69141) (24, 2.68359) (32, 2.64258) (48, 2.62695) (64, 2.67383) (96, 2.70898) (128, 2.69531) (192, 2.67188) (256, 2.71094) (384, 2.70117) (500, 2.72461)
}
\newcommand{\TimeStepsEvalMambaMSE}{
(2, 49.96407) (4, 0.31226) (6, 0.03766) (8, 0.02965) (12, 0.02825) (16, 0.02895) (24, 0.02982) (32, 0.02981) (48, 0.03076) (64, 0.03153) (96, 0.03013) (128, 0.03087) (192, 0.03120) (256, 0.03412) (384, 0.03320) (500, 0.03424)
}
\newcommand{\TimeStepsEvalMambaVel}{
(2, 100) (4, 100) (6, 3.72070) (8, 2.94922) (12, 2.69141) (16, 2.64453) (24, 2.62891) (32, 2.63672) (48, 2.63477) (64, 2.68750) (96, 2.68164) (128, 2.60156) (192, 2.68945) (256, 2.70898) (384, 2.69727) (500, 2.71094)
}
\newcommand{\TimeStepsEvalMambaSwa}{
(2, 100) (4, 8.64844) (6, 2.98242) (8, 2.57422) (12, 2.45312) (16, 2.43555) (24, 2.48633) (32, 2.47266) (48, 2.48242) (64, 2.56055) (96, 2.57422) (128, 2.61133) (192, 2.57617) (256, 2.56836) (384, 2.57617) (500, 2.55469)
}
\definecolor{C0}{rgb}{0.121569, 0.466667, 0.705882}
\definecolor{C1}{rgb}{1.000000, 0.498039, 0.054902}
\definecolor{C2}{rgb}{0.172549, 0.627451, 0.172549}
\definecolor{C3}{rgb}{0.839216, 0.152941, 0.156863}
\definecolor{C4}{rgb}{0.580392, 0.403922, 0.741176}
\definecolor{C5}{rgb}{0.549020, 0.337255, 0.294118}
\definecolor{C6}{rgb}{0.890196, 0.466667, 0.760784}
\definecolor{C7}{rgb}{0.498039, 0.498039, 0.498039}
\definecolor{C8}{rgb}{0.737255, 0.741176, 0.133333}
\definecolor{C9}{rgb}{0.090196, 0.745098, 0.811765}

\newcommand{\figTimeSteps}{
\begin{figure}[t]
\centering

\begin{subfigure}[b]{0.49\columnwidth}
\begin{tikzpicture}
\begin{axis}[
  width=\textwidth, height=0.7\textwidth,
  xmin={4}, xmax={500}, xmode={log}, xtick={8, 32, 128,500}, xticklabels={8, 32, 128,500},
  ymin={0}, ymax={0.35},
  grid={major},
]
\addplot[C0] coordinates {\TimeStepsEvalSFourMSE};
\addplot[C1] coordinates {\TimeStepsEvalMambaMSE};
\legend{
  {S4},
  {Mamba}
}
\end{axis}
\end{tikzpicture}
\caption{\mse{}}
\end{subfigure}
\hfill
\begin{subfigure}[b]{0.49\columnwidth}
\begin{tikzpicture}
\begin{axis}[
  width=\textwidth,height=0.7\textwidth,
  xmin={4}, xmax={500}, xmode={log}, xtick={8, 32, 128,500}, xticklabels={8, $32$, $128$,$500$},
  ymin={2}, ymax={5.0},
  grid={major},
]
\addplot[C0] coordinates {\TimeStepsEvalSFourVel};
\addplot[C1] coordinates {\TimeStepsEvalMambaVel};
\addplot[dashed, C0] coordinates {\TimeStepsEvalSFourSwa};
\addplot[dashed, C1] coordinates {\TimeStepsEvalMambaSwa};
\legend{
  {S4 Speed},
  {Mamba Speed},
  {S4 swa},
  {Mamba swa}
}
\end{axis}
\end{tikzpicture}
\caption{$MSE$ speed}
\end{subfigure}

\caption{Ablation study of the number of reverse diffusion steps for our two best models, one with S4 and the other with Mamba. We evaluate them with all four metrics: (a) the sings score, (b) the \mse{} and, (c \& d) the $MSE$ between the generated and the true next window for the speed and the swa, respectively.}
\label{fig:TimeSteps}
\end{figure}
}

\newcommand{\figTimeStepss}{
\begin{figure}[t]
\centering

\begin{subfigure}[b]{0.49\columnwidth}
\centering
         \includegraphics[width=\textwidth]{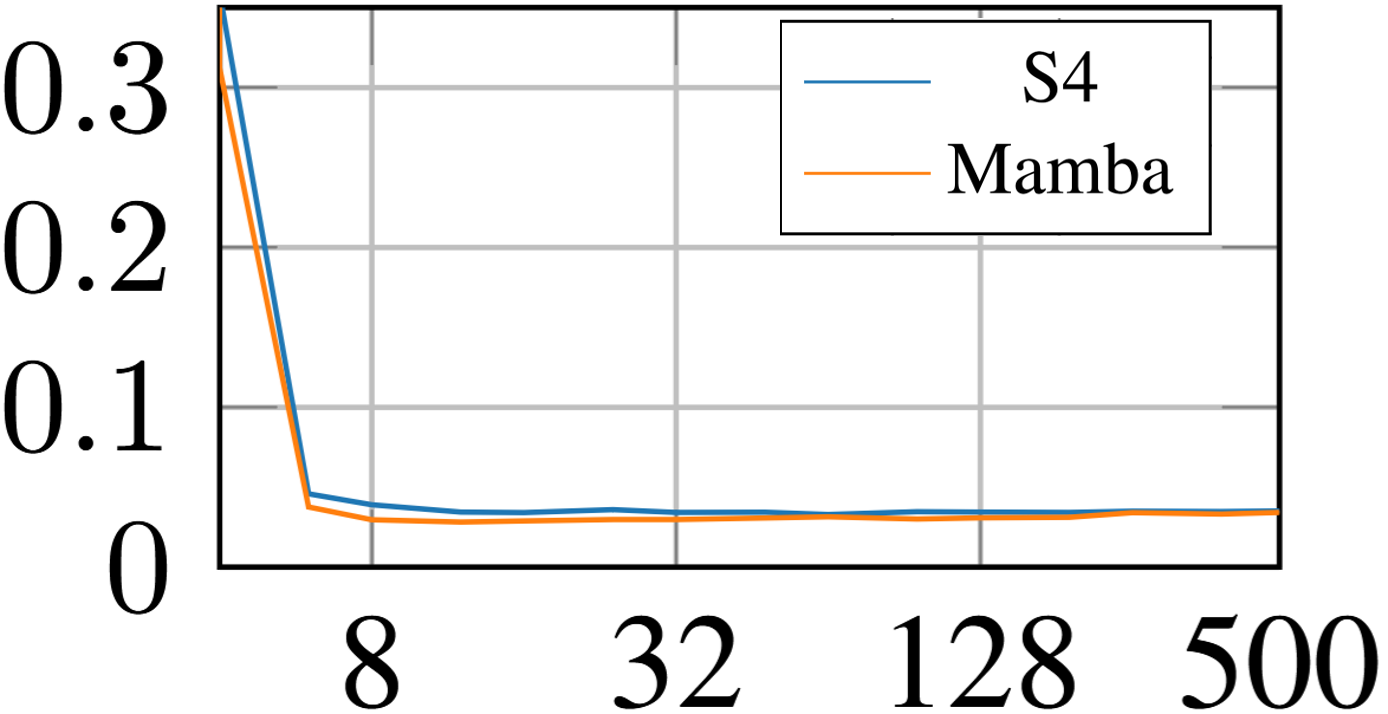}
         \caption{\mse{}}
\end{subfigure}
\hfill
\begin{subfigure}[b]{0.49\columnwidth}
\centering
         \includegraphics[width=\textwidth]{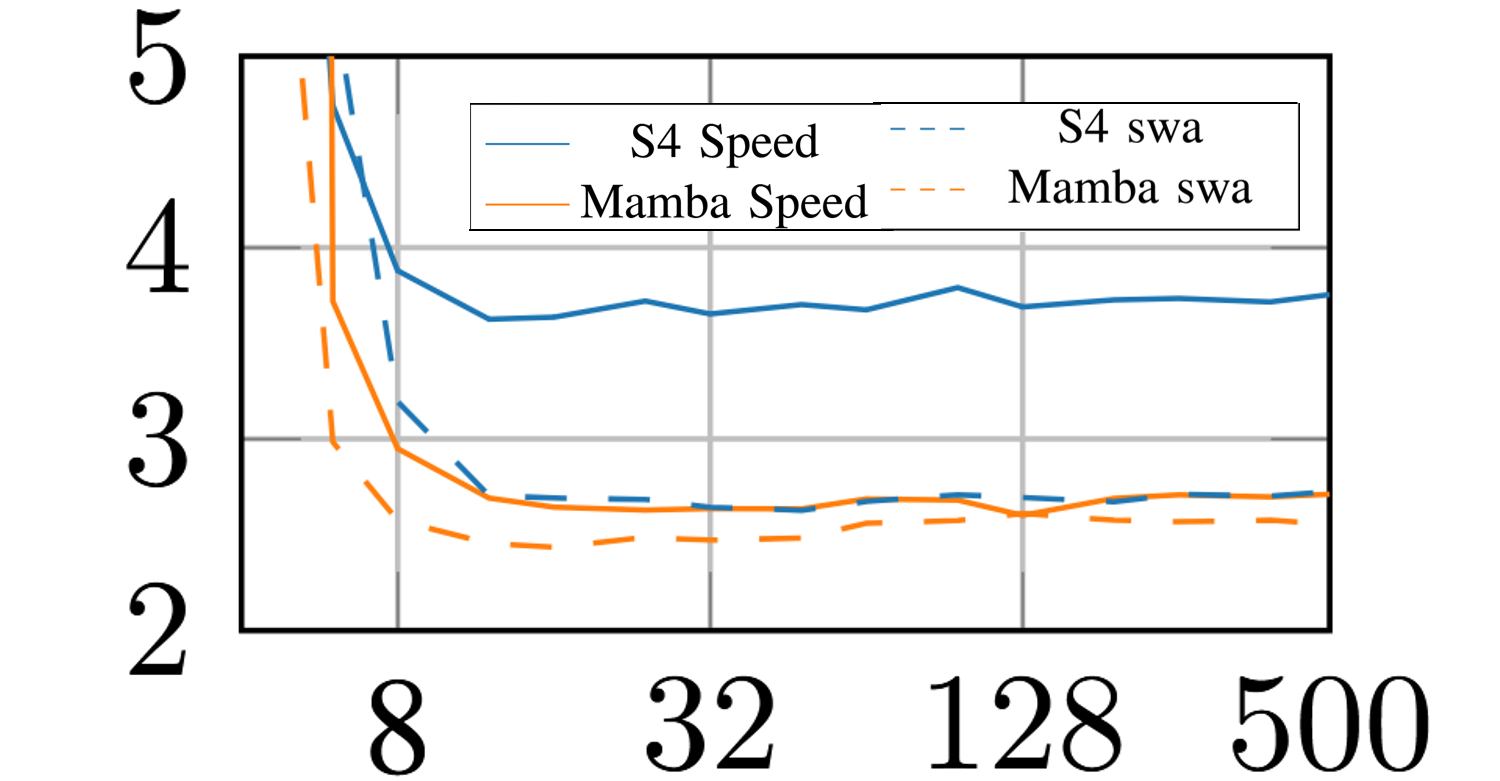}
         \caption{MSE for speed and swa}
\end{subfigure}
\caption{Ablation study of the number of reverse diffusion steps for generated windows of SSSD\textsuperscript{S4} and SSSD\textsuperscript{Mamba}.}
\label{fig:TimeSteps}
\end{figure}
}

\section{Introduction}
Time series are ubiquitous in almost  every part in our daily lives in form of 
climate measurements, recorded speech, econometrics, mathematical finance, or healthcare.
However, real-world datasets can be challenging to work with. For example, datasets that are too small can limit the application and generalization capabilities of machine learning-based approaches, or incorrect time series (e.g., due to missing values or outliers) are likely to be discarded, making the dataset even smaller. 
Generative models can remedy these problems through synthetic data generation or augmentation \cite{b1}, or by imputing missing values in time series \cite{b2}. In turn, they can help improve downstream tasks, such as classification, anomaly detection, or motion planning and control in intelligent transportation systems (ITS)  \cite{b1, b3}.

Related work on time series generation can be categorized into autoregressive (AR) and non-autoregressive (non-AR). AR models iteratively generate new datapoints based on their own predictions. Transformer-based architectures \cite{b4} became a common choice over CNNs or RNNs due to better long-term dependency modeling. However, their quadratic runtime poses challenges for long time series and computationally constrained settings, leading to recent works focusing on improving both runtime and prediction quality \cite{b5}. More recently, state space models (SSMs), such as S4 \cite{b6} and Mamba \cite{b7} challenge the dominance of transformers and CNNs in AR time series generation. For non-AR generation, VAE and GAN-based approaches offer fast inference but can struggle with complex temporal dynamics or training challenges such as mode-collapse or vanishing gradients. Recent advances in diffusion models outperform VAEs \cite{b8} and GANs \cite{b9} in terms of generated data quality and training stability, but this comes at the cost of longer sampling times. Despite diffusion models showing strong performance for time series generation in healthcare \cite{b2, b10} or recommender systems \cite{b11}, their application to generation or imputation of automotive time series data (e.g., vehicle CAN data) remains underexplored. This leads to the question: \textit{Can diffusion models effectively learn to generate and impute synthetic vehicle CAN data from a limited number of training samples?}

In this paper, we answer this question by proposing a new approach that applies diffusion models to both generation and imputation of vehicle CAN-data. More specifically, we are given a dataset with a limited number of long-term vehicle CAN-data recordings of different car models that drive the same pre-defined test track. Our goal is to generate new realistic laps, with realistic physics and a plausible driving behavior. To this end, we propose a hybrid generative approach combining AR and non-AR techniques, where we sequentially generate windows in an AR manner while using a non-AR setting to generate each window. This increases the number of training samples while reducing the size of the generated sample, making training feasible.

We evaluate the architectures of recent works that have shown to work well with diffusion models for time series generation, i.e., an efficient SSM approximation called LongConv \cite{b12} and SSSD\textsuperscript{S4} \cite{b2}, which uses the S4 SSM framework \cite{b6}. With these models and all available information from the dataset, we systematically explore different network input configurations and incorporate domain knowledge that helps enhance both training and sampling. Finally, we introduce a novel SSSD variant where S4 layers are replaced with the recently proposed Mamba \cite{b7} layers. 
We evaluate all of our approaches on a real-world CAN dataset. It has a limited number of 66 samples, each with a significant length of 12,554 data points, which makes it challenging to train a generative model. To evaluate the quality of the generated samples we propose three different measures to verify the physical correctness and the driving behavior. They capture the relationship between torque and acceleration, and quantify the adherence to the test track. Furthermore, the dataset contains physically implausible regions, which we correct using our best model in a real-world setting to improve the quality of the data. We evaluate two imputation approaches: a naive approach and an adapted technique from the image domain to better harmonize imputed regions with the surrounding data, called RePaint \cite{b13}.

We find that diffusion models can effectively generate and impute vehicle CAN-data even when trained on a challenging long-range time series dataset with a limited number of samples. Our novel SSSD variant, together with our dataset specific improvements, generates samples which even exceed the training data in terms of physical correctness, while at the same time showing plausible driving behavior, both evaluated by our proposed measures. The same model can be used in both evaluated imputation approaches to correct physically implausible regions in the training data. RePaint yields slightly worse physics than the naive approach, but harmonizes better with the rest of the data. Our main contributions are: 
\begin{itemize}
    \item We apply diffusion models to a real-world vehicle CAN dataset with a limited number of samples of significant length. Therefore, we propose a hybrid generation technique to overcome the challenges of the dataset.
    \item We evaluate and enhance different network architectures proposed in the recent diffusion model and time series literature as well as different network input configurations.
    \item  We propose three different measures for evaluating the generated samples in terms of physical correctness and driving behavior.
    \item We evaluate the imputation capabilities of our best model on physically implausible regions in the training data with two different approaches, which demonstrates that we can improve the physical correctness of the dataset.
\end{itemize}

\section{Approach}
Generating synthetic vehicle CAN data using diffusion models can serve multiple applications in the automotive industry. For example, it can expand existing datasets for potential downstream tasks, such as motion planning and control in ITS, as well as accelerate new vehicle development by providing simulated sensor data. In addition, it can allow imputation of missing values in existing datasets and correction of readings from faulty sensors without explicit training. In this section, we describe our approach and network architecture for generating and imputing realistic sensor data that closely approximates real-world driving.
\subsection{Non-autoregressive Training and Window Generation}
\label{sec:training}
\figApproach
We are given a dataset $\mathcal{D}$ of sensor readings collected from multiple vehicles traversing the same track over repeated iterations, which we define as ``laps''. Each vehicle is equipped with four sensors that capture the speed, the two torques at the drive shafts and the steering wheel angle. Unlike most time series, the points in our dataset are sampled every half meter, so that each lap has the same length $T$. For each lap $l \in \mathcal{D}$, we denote the multivariate time series as $X^l = \{x_t^l\}_{t=1}^{T}$, where $x_t^l \in \mathbb{R}^4$ represents the four-dimensional sensor reading at position $t$. We pad each lap with 512 data points at the beginning and segment each lap's time series into overlapping windows $W_i^l$ of size $w = 1024$, with a stride of $s$, so that $W_i^l = \{x_t^l\}_{t=i}^{i+w-1}$, with $i\in[0,s,2s,...]$. Each window $W_i^l$ is further divided into two equal halves: $W_i^l = [W_{i,\text{past}}^l, W_{i,\text{future}}^l]$, where $W_{i,\text{past}}^l = \{x_t^l\}_{t=i}^{i+w/2-1}$ and $W_{i,\text{future}}^l = \{x_t^l\}_{t=i+w/2}^{i+w-1}$. Similarly, additional contextual features $C^l = \{c_t^l\}_{t=1}^{T}$, such as the track elevation profile, are processed into matching overlapping windows $C_i^l = \{c_t^l\}_{t=i}^{i+w-1}$ with the same stride $s$.

For training our forecasting diffusion model $f_\theta$, we apply a noise corruption process to $W_{i,\text{future}}^l$ according to a diffusion schedule. Specifically, at timestep $n \in \{1,2,...,N\}$, we sample noise $\epsilon \sim \mathcal{N}(0, I)$ and corrupt the future window as:
\begin{equation}W_{i,\text{future}}^{l,(n)} = \sqrt{\bar{\alpha}_n}W_{i,\text{future}}^l + \sqrt{1-\bar{\alpha}_n}\epsilon\end{equation}
where $\{\bar{\alpha}_n\}_{n=1}^N$ is defined by the noise schedule.

$f_\theta$ is trained to predict the noise component $\epsilon$ from the corrupted future window, from $W^{l,(n)}_{i,\text{input}} = [W_{i,\text{past}}^l, W_{i,\text{future}}^{l,(n)}]$ and the contextual features $C_i^l$. The loss function is computed exclusively over the future half of the window:
\begin{equation}\mathcal{L}(\theta) = \mathbb{E}_{l,i,n,\epsilon}\left[ \| \epsilon - f_\theta(W_{i,\text{input}}^{l,(n)}, C_i^l, n) \|_2^2 \right]\end{equation}
The segmentation and training process is shown in Fig. \ref{fig:approach}. The segmentation into overlapping windows increases the number of training samples and reduces the complexity of the generation task, since the generated samples are shorter.

For our model, we evaluate two neural network architectures, LongConv \cite{b12} and SSSD\textsuperscript{S4} \cite{b2}. For SSSD\textsuperscript{S4},  we add layers for the conditioning input, so that all single values are processed with a fully connected layer, while a 1D convolution handles the time series, since unlike \cite{b2}, our conditioning contains time-series and single values. The two inputs are then added and fed into another convolution layer, before being passed to the residual blocks. We also compare the vanilla SSSD\textsuperscript{S4} with a variant where S4 layers are replaced with Mamba \cite{b7}, called SSSD\textsuperscript{Mamba}.

\subsection{Conditional Autoregressive Lap Generation}
To generate a complete synthetic lap, we autoregressively generate multiple candidate  future windows based on the last generated window, summarized in Algorithm \ref{alg:gen}. We start by randomly drawing a padded window $W_{i,\text{past}}\in \mathbb{R}^{512 \times 4}$ from the beginning of a training lap. During the generation process, we sample $b=16$ candidate future windows $\{\hat{W}_{i,\text{future}}^{(j)}\}_{j=1}^{b}$ using the reverse diffusion process:
\begin{equation}
\hat{W}_{i,\text{future}}^{(j)} = \mathcal{G}_\theta(W_{i,\text{past}}, C_i, j)
\end{equation}
where $\mathcal{G}_\theta$ represents the generative process of our trained diffusion model. We find 16 candidate windows as a good tradeoff between quality and performance. We then evaluate each candidate window using our proposed quality metric $\mathcal{M}$ (presented in Section \ref{sec:physic_eval}), which evaluates the physical plausibility, and select the optimal window index:
\begin{equation}j^* = \arg\min_{j \in \{1,2,...,b\}} \mathcal{M}(\hat{W}_{i,\text{future}}^{(j)})\end{equation}
The selected optimal window $\hat{W}_{i,\text{future}}^{(j^*)}$ is then used as the next past window to generate the next future window. This autoregressive procedure continues until a complete lap is synthesized. The multi-candidate sampling approach with quality-based selection mitigates error accumulation and enhances the overall coherence and physical plausibility of the generated lap. This approach effectively balances between maintaining the stochastic nature of the diffusion model while ensuring the generated trajectory adheres to the physical constraints of the vehicle and track.
\MyAlgorithm

\subsection{Further Application: Imputation}
\label{sec:app-impuation}
Many real-world datasets contain missing or erroneous values, which may require discarding samples, resulting in smaller and less diverse datasets. Imputation of these values can remedy this problem.

In general, the application of our model, trained in Section \ref{sec:training}, to imputation is straightforward. We denote the ground truth future windows as  $W_{i,\text{gt}}^ {l}$, the unknown data points as $m \odot W_{i,\text{gt}}^ {l}$ and the known data points as $(1 - m) \odot W_{i,\text{gt}}^ {l}$. Since each reverse step from $W_{i,\text{future}}^ {l, (n)}$ to $W_{i,\text{future}}^ {l, (n-1)}$ depends only on $W_{i,\text{future}}^ {l, (n)}$, we can modify the known regions $(1 - m) \odot W_{i,\text{gt}}^ {l}$ while maintaining the distributional properties. The forward process's Markov Chain of Gaussian noise allows sampling the intermediate time series $W_{i,\text{gt}}^ {l,(n)}$ at any time step. Thus, we can sample known regions $m \odot W_{i,\text{gt}}^ {l,(n)}$ at any time $n$. Combining the unknown regions and the known regions, one reverse step with $W^{l,(n)}_{i,\text{input}} = [W_{i,\text{past}}^l, W_{i,\text{future}}^{l,(n)}]$ becomes:
\begin{subequations}
\begin{align}
  W_{i,\text{gt}}^ {l, (n-1)} &\sim \mathcal{N}(\sqrt{\bar{\alpha}_t} W_{i,\text{gt}}^ {l}, (1-\bar{\alpha}_t) \mathbf{I}) \\
  W_{i,\text{unknown}}^ {l, (n-1)} &\sim \mathcal{N}(\mu_{\theta}(W_{i,\text{input}}^ {l, (n)}, t), \Sigma_{\theta}(W_{i,\text{input}}^ {l, (n)}, t)) \\
   W_{i,\text{future}}^ {l, (n-1)} &= m \odot W_{i,\text{gt}}^ {l, (n-1)} + (1-m) \odot W_{i,\text{unknown}}^ {l, (n-1)}
\end{align} \label{eq:ourStep}
\end{subequations}
Here, $ W_{i,\text{gt}}^ {l, (n-1)}$ is sampled using known data points from $m \odot  W_{i,\text{gt}}^ {l}$, while $W_{i,\text{unknown}}^ {l, (n-1)}$ is sampled from the model conditioned on $W_{i,\text{future}}^ {l, (n)}$ from the previous iteration.

In our experiments, we experiment with two different time schedules. First, a naive time schedule simply proceeds from the chosen number of reverse steps to 0. This often results in generated content that matches the structure but fails semantically at the boundaries between known and unknown regions. The problem arises because the model generates the entire sample, but the known region gets replaced by the noisy known data points without considering the generated parts of the sample, leading to disharmony. Second, to solve this issue, we adapt the advanced time schedule of RePaint \cite{b13}. Intermediate samples $W_{i,\text{future}}^ {l, (n)}$ are diffused back $j$ time steps to $W_{i,\text{future}}^ {l, (n+j)}$ using the forward process before proceeding with the next reverse steps. This operation is repeated $r$ times and allows the generated regions to better harmonize with known regions.

\section{Experimental Setup}
In this section, we present the utilized dataset together with additional inputs that we use in our experiments in Section \ref{sec:dataset}, our evaluation measures in Section \ref{sec:physic_eval} and \ref{sec:tam}, and the setup for our imputation experiment in Section \ref{sec:exp_imputation}.

\subsection{Dataset and Additional Model Inputs}
\label{sec:dataset}
Our evaluation uses a confidential automotive dataset containing sensor recordings from eight vehicles, traversing an identical track over repeated
iterations, with a total of 66 “laps”. Each lap has 12554 samples and, unlike conventional time series, they are recorded at 0.5-meter intervals instead of fixed time intervals, yielding consistent sample counts across laps. Each sample contains four channels: speed, left/right driveshaft torque, and steering wheel angle (swa). 

In addition to the sensor data, the dataset contains other information that we leverage in our experiments to generate realistic laps. This includes the track \textit{elevation} profile, which we standardize, and the \textit{vehicle parameters}, i.e., the mass, the cW\textsubscript{A}-value (wind resistance), and the wheel perimeter. For the model to know where on the track to generate a window, we number each data point from 0 to 12553, embed them using sinusoidal embeddings \cite{b4}, and input them as the \textit{track position}.

To generate the first window of a track, we need to pad the training data at the beginning to have a window to condition the model on. We empirically evaluated different padding strategies. Zero padding causes jumps as the laps start with moving vehicles, while same padding would create physically inconsistent states with no acceleration but high torques coming from the initial acceleration phase of the lap. To mitigate these issues, we implemented \textit{physically correct same padding} (PCSP). PCSP uses same-padding for speed and steering wheel angle, while computing appropriate torques for constant speed. We add some noise to all channels for realism. To account for the padded values, we implement a binary \textit{padding indicator} to mark padded points.
\subsection{Physical Evaluation Measure}
\label{sec:physic_eval}

A generated lap should be physically correct and it should follow the original track. We propose a metric comparing the measured car acceleration against the predicted acceleration based on torques and car properties, with smaller differences indicating higher correctness.

Due to sensor noise, we first smooth the data using a linear filter and then compute the acceleration as $a_v = \frac{v_{t+2}-v_t}{2 \Delta t}$. Since data is sampled by space (every half meter) rather than time (see Section \ref{sec:dataset}), $\Delta t$ varies per sample and can be approximated by $\Delta t \approx \big(\dfrac{v_{t+2}+v_t}{2}\big)^{-1}$ seconds. We compare $a_v$ with the predicted acceleration $a_{torque}$ calculated via Newton's second axiom: $a_{torque} = (F_{torques} - F_{downhill} -F_{wind} - F_{rolling}) / m$, using the torque force, the downhill force, the wind and rolling resistance, and the car mass $m$.

As the torque sensors do not capture braking, $a_{torque}$ is reliable only during non-braking periods, identified when the measured torques are negative. For these segments, we compute \msefull{} as the mean squared error between $a_v$ and $a_{torque}$. To mitigate outlier effects, caused by windows containing almost only breaking behavior, we discard 5\% of the windows with worst \msefull{} (denoted as \mse). To also capture the behavior when breaking, we calculate a \textit{signs score} representing the percentage of data points where the sings of $a_v$ and $a_{torque}$ match.

Validation on training data shows that most laps achieve an \msefull{} below 0.02 and a mean signs score of 0.91, confirming that our measures effectively assess physical plausibility. However, miscalibrated torque sensors create outliers in the dataset (see Fig. \ref{fig:MSEgood}), resulting in an overall training data \msefull{} of 0.084.

\figMSEgood

\subsection{Trajectory Adherence}
\label{sec:tam}

Besides the physics, a generated lap must reflect a realistic driving behavior where the car remains on track. To quantify this realism, we propose the Trajectory Adherence Metric (TAM), which compares the generated data against the training data distributions along the track. For each car, we establish minimum and maximum values for speed and swa at each track position, creating a "realistic" range. We evaluate the generated data point-by-point, assigning 0 if values fall within the range, or adding the absolute deviation to a running total if outside of this range. The final score is the mean over the track length, representing the mean deviation from realistic behavior. Lower scores indicate greater similarity to the training data. Fig. \ref{fig:LabGen} visualizes the range of realistic values.

\subsection{Imputation}
\label{sec:exp_imputation}
For imputing physically implausible regions we first have to identify regions with a high \msefull. We consider data points with a \msefull{} greater than $0.2$ to be implausible and combine them into regions. We discard regions with less than 70 data points, as they are often artifacts of smoothing during the \msefull{} computation, enlarge the regions by 72, and combine regions that are less than 500 data points apart to reduce the harmonization problem (see Section \ref{sec:app-impuation}). For each track, we set the first implausible region at the beginning of the generated window and impute all implausible data points. We continue with the next region until all regions are imputed. As for the generation of a lap, we impute each window 16 times and take the one with the lowest \msefull. During our experiments, we want to evaluate the effect of the two time schedules described in Section \ref{sec:app-impuation}, the number of reverse diffusion steps, and the difference between imputing all channels or only the two torque channels.
To evaluate the results, we compute the \msefull{} for the imputed dataset and quantify the continuity at the boundaries of the imputed regions for each channel and for the start and end of a region separately. For each imputed region, we compute the absolute difference between the observed discontinuity in the imputed series and the corresponding difference in the original data, e.g., for a boundary at the beginning of a region at position $t$, we compute $\lvert\lvert imputed[t] - imputed[t-1]\rvert - \lvert original[t] - original[t-1]\rvert\rvert$ and average over all regions. This metric directly assesses how well the imputed region harmonizes with the rest of the data, taking into account the necessary changes due to the track, e.g., when there is a curve, a larger difference in the swa is realistic. We expect the average discontinuity at the beginning of the regions to be lower than at the end, since the model is trained to generate data after the previous window.

\tabCar
\vspace{-0.5em}
\tabCarLap
\section{Results}
This section provides an overview of the
implementation details, followed by the results of our experiments.

\subsection{Implementation Details}
In our experiments, we use our model to forecast a window based on the previous one. For additional information, we concatenate continuous inputs described in Section \ref{sec:dataset} channel-wise to the input. We use MSE as our loss function, an SGD optimizer with a learning rate of 0.0004 and a batch size of 32 for 100 epochs, resulting in a total training time of 3 hours on an NVIDIA A100 GPU. To evaluate the model, we generate 4096 windows and 16 laps per model with $N=500$ time steps. For the imputation with RePaint, we set $j=r=5$. Note that there will always be some variation in the results due to the stochastic nature of the training and sampling process of diffusion models.

\subsection{Results}
In this section, we present our generation experiments using the three network architectures, LongConv, SSSD\textsuperscript{S4}, and SSSD\textsuperscript{Mamba}, including their network input configuration. All results are summarized in Table \ref{tab:Car}.

\textbf{LongConv} In our first experiment, we use the LongConv architecture and input the zero-padded sensor data, the elevation, the track position embedding, and the vehicle parameters into the network. Since LongConv has no dedicated input for non-time series data, such as the vehicle parameters, we repeat them and stack them on top of the time series data. 

When generating only a single next window, LongConv is already able to outperform the training data in terms of the \msefull. However, there is a small but noticeable discontinuity between the windows and the difference in speed and swa to the original next window is relatively large. When generating a lap, this causes the model to diverge, after the generation a few windows.

\textbf{SSSD\textsuperscript{S4}} In our next experiment, we use SSSD\textsuperscript{S4} as our network architecture. We feed all additional inputs, e.g., the elevation, the vehicle data and the padding indicator into our improved version of SSSD's condition input. We also use our proposed PCSP.

The use of SSSD\textsuperscript{S4} reduces the \msefull{} to 0.036 and the difference for the generated speed and swa in comparison to the original next window enabling the model to generate a full lap without diverging. The generated laps outperform the training data by far in terms of physical plausibility, evaluated by both the signs score and the \msefull.  However, the speed and the swa differs significantly from the realistic range of values, exhibiting an unrealistic driving behavior where the car would not stay on track. 

\imputation

For the SSSD\textsuperscript{S4} experiment, we perform an ablation to study the effect of the network input configuration. We run this ablation experiment with zero-padded training laps instead of PCSP, the original architecture in the conditioning branch, and without the padding indicator. This experiment studies the effect of the input configuration and allows a direct comparison with LongConv. The results are, except for the signs score, worse than the results of the SSSD\textsuperscript{S4} experiment, but significantly better than the results of LongConv, indicating that both, the different network architecture and the network input configuration improved the results.
\figGenTrack

\textbf{SSSD\textsuperscript{Mamba}} In our last experiment we use the architecture where we replaced the S4 layers with Mamba layers, but keep the input configuration the same.

For only generating the next window, only the difference to the original window for the speed and swa decreases, but the physics stay nearly unchanged. Generated laps demonstrate improved trajectory adherence with minimal physical accuracy loss. Since the qualitative analysis shows that the speed and the swa deviate only slightly from the realistic range, we consider the generated laps to have a plausible driving behavior. See Fig. \ref{fig:LabGen} for an example.

\figTimeStepss

\textbf{Ablation Diffusion Steps} We perform an ablation for our best models on the impact of the number of reverse diffusion steps. Interestingly, the results, shown in Fig. \ref{fig:TimeSteps}, show that the best physical results are obtained with only 12 steps, while the best MSE speed/swa are generated with 16 steps.

\subsection{Imputation}
In this section, we evaluate the capabilities of our SSSD\textsuperscript{Mamba} model to impute and correct physically implausible regions, a task for which the model was not trained for. We test a naive and an advanced schedule for the time steps in the reverse diffusion process described in Section \ref{sec:app-impuation}. 

Table \ref{tab:Imputation} shows that all imputation methods improved the physical plausibility of the dataset, reducing \msefull{} from 0.084 to 0.041 in the best case. The advanced schedule yielded worse physical plausibility, particularly when imputing all channels, although it produced fewer boundary discontinuities except in the speed channel, supporting \cite{b13}'s findings. For the advanced schedule, generating only torques leads to less discontinuities in the torque channel, while there are naturally no discontinuities in the other channels. For the naive schedule, there is no such significant difference in the torque channel. Generally, increasing reverse diffusion steps reduced discontinuities across experiments.

\section{Conclusion}
We presented a hybrid generative approach that combines autoregressive and non-autoregressive techniques. It enabled us to train a diffusion model on a challenging CAN dataset with limited lengthy samples and to impute physically implausible regions, which allows future work to extend and correct the dataset for potential downstream tasks. We evaluated two recent time series generation architectures alongside our novel SSSD\textsuperscript{Mamba} variant. With optimized configurations, our generated samples surpassed the training data in physical plausibility while maintaining realistic driving behavior. Our proposed metrics confirmed these improvements. When imputing physically implausible regions, we improved the physics of the training data, although boundary discontinuities persisted with both schedules tested, they appear less frequently with RePaint. Future work has to show how our approach generalizes to more diverse and complex datasets and vehicle types.


\begin{thebibliography}{00}
 \bibitem{b1} B. Trabucco, K. Doherty, M. A. Gurinas, and R. Salakhutdinov, “Effective data augmentation with diffusion models,” in The Twelfth International Conference on Learning Representations, 2023.
 \bibitem{b2} J. L. Alcaraz and N. Strodthoff, “Diffusion-based time series imputation and forecasting with structured state space models,” Transactions on Machine Learning Research, 2022.
 \bibitem{b3} J. Wyatt, A. Leach, S. M. Schmon, and C. G. Willcocks, “Anoddpm: Anomaly detection with denoising diffusion probabilistic models using simplex noise,” in 2022 IEEE/CVF Conference on Computer Vision and Pattern Recognition Workshops (CVPRW), IEEE Computer Society, 2022, pp. 649–655.
 \bibitem{b4} A. Vaswani, N. Shazeer, N. Parmar, J. Uszkoreit, L. Jones, et al., “Attention is all you need,” Advances in neural information processing systems, vol. 30, 2017.
 \bibitem{b5} H. Zhou, S. Zhang, J. Peng, S. Zhang, J. Li, et al., “Informer: Beyond efficient transformer for long sequence time-series forecasting,” in Proceedings of the AAAI conference on artificial intelligence, vol. 35, 2021, pp. 11 106–11 115.
 \bibitem{b6} A. Gu, K. Goel, and C. Re, “Efficiently modeling long sequences with structured state spaces,” in International Conference on Learning Representations, 2022.
 \bibitem{b7} A. Gu and T. Dao, Mamba: Linear-time sequence modeling with selective state spaces, 2024. [Online]. Available: https://openreview.net/forum? id=AL1fq05o7H.
 \bibitem{b8} D. P. Kingma and M. Welling, “Auto-encoding variational bayes,” stat, vol. 1050, p. 1, 2014.
 \bibitem{b9} I. Goodfellow, J. Pouget-Abadie, M. Mirza, B. Xu, D. Warde-Farley, et al., “Generative adversarial nets,” Advances in neural information processing systems, vol. 27, 2014.
 \bibitem{b10} B. Aristimunha, R. Y. de Camargo, S. Chevallier, O. Lucena, A. G. Thomas, et al., “Synthetic sleep eeg signal generation using latent diffusion models,” in Deep Generative Models for Health Workshop NeurIPS 2023, 2023.
 \bibitem{b11} Z. Yang, J. Wu, Z. Wang, X. Wang, Y. Yuan, et al., “Generate what you prefer: Reshaping sequential recommendation via guided diffusion,” in Thirty-seventh Conference on Neural Information Processing Systems, 2023.
 \bibitem{b12} J. Vetter, J. H. Macke, and R. Gao, “Generating realistic neurophysiological time series with denoising diffusion probabilistic models,” bioRxiv, 2023.
 \bibitem{b13} A. Lugmayr, M. Danelljan, A. Romero, F. Yu, R. Timofte, et al., “Repaint: Inpainting using denoising diffusion probabilistic models,” in Proceedings of the IEEE/CVF conference on computer vision and pattern recognition, 2022, pp. 11 461–11 471.
\end{thebibliography}
\end{document}